\documentclass[conference, twoside]{IEEEtran}
\IEEEoverridecommandlockouts
\usepackage{amsmath,graphicx}
\usepackage{multirow}
\usepackage{amssymb}
\usepackage{algorithm}
\usepackage{nohyperref}
\usepackage{url}
\usepackage{epstopdf}
\usepackage{amsthm}
\usepackage{enumitem}
\usepackage{cite}
\usepackage[numbers]{natbib}
\usepackage{soul}
\usepackage{subfig}
\usepackage{times}
\usepackage{makecell}
\usepackage[noend]{algorithmic}
\usepackage{color,soul}
\soulregister\cite7
\soulregister\ref7
\soulregister\pageref7
\usepackage{diagbox}
\usepackage[export]{adjustbox}
\renewcommand{\footnotesize}{\scriptsize}
\usepackage{bm}

\newtheorem{mydef}{Definition}

\usepackage{soul}
\usepackage{comment}

\usepackage{fancyhdr}

\fancypagestyle{specialfooter}{%
  \fancyhf{}
  \fancyhead[CO]{IJCNN 2020. International Joint Conference on Neural Networks. Glasgow, UK. 19-24 July 2020}
  
  \fancyfoot[C]{Personal use is permitted, but republication/distribution requires IEEE permission.\\
See http://www.ieee.org/publications standards/publications/rights/index.html for more information.}
}

\fancypagestyle{mypagestyle}{%
  \fancyhf{}% Clear header/footer
  \fancyhead[CE]{IJCNN 2020. International Joint Conference on Neural Networks. Glasgow, UK. 19-24 July 2020}% Author on Odd page, Centred
  \fancyhead[CO]{Ontology-based Interpretable Machine Learning for Textual Data}% Title on Even page, Centred
  \fancyfoot[C]{--~\thepage~--}%
}
\usepackage{cite}
\usepackage{amsmath,amssymb,amsfonts}
\usepackage{algorithmic}
\usepackage{graphicx}
\usepackage{textcomp}
\usepackage{xcolor}
\def\BibTeX{{\rm B\kern-.05em{\sc i\kern-.025em b}\kern-.08em
    T\kern-.1667em\lower.7ex\hbox{E}\kern-.125emX}}
    
\begin{document}

\title{Ontology-based Interpretable Machine Learning \\for Textual Data
% \thanks{978-1-7281-2009-6/$ \$31.00$ \copyright2020 IEEE}
}

\author{\IEEEauthorblockN{Phung Lai, NhatHai Phan, Han Hu, Anuja Badeti}
\IEEEauthorblockA{
\textit{New Jersey Institute of Technology, USA}\\
\{tl353,phan,hh255,ab2253\}@njit.edu}
\and
\IEEEauthorblockN{David Newman}
\IEEEauthorblockA{
\textit{Wells Fargo Bank, USA}\\
David.Newman@wellsfargo.com}
\and
\IEEEauthorblockN{Dejing Dou}
\IEEEauthorblockA{
\textit{University of Oregon, USA}\\
dou@cs.uoregon.edu}
}

% \markboth{{IEEE IJCNN 2020. International Joint Conference on Neural Networks. Glasgow, UK. 19-24 July 2020}}{{Ontology-based Interpretable Machine Learning for Textual Data}}

% \pagestyle{headings}
% % The paper headers

\maketitle
\thispagestyle{specialfooter}

% \cfoot{2020 International Joint Conference on Neural Networks (IJCNN)} 
\begin{abstract}
 In this paper, we introduce a novel interpreting framework that learns an interpretable model based on an ontology-based sampling technique to explain agnostic prediction models. Different from existing approaches, our algorithm considers contextual correlation among words, described in domain knowledge ontologies, to generate semantic explanations. To narrow down the search space for explanations, which is a major problem of long and complicated text data, we design a learnable anchor algorithm, to better extract explanations locally. A set of regulations is further introduced, regarding combining learned interpretable representations with anchors to generate comprehensible semantic explanations. An extensive experiment conducted on two real-world datasets shows that our approach generates more precise and insightful explanations compared with baseline approaches.
\end{abstract}
\IEEEpubid{0000--0000/00\$00.00˜\copyright˜2015 IEEE
}

\section{Introduction}

In critical scenarios, such as clinical practices, having the ability to interpret machine learning (\textbf{ML}) 
model outcomes is significant to reduce the error rate and improve the trustworthiness of ML-based systems \citep{robnik2008explaining, goyal2017making}. To achieve this, typical approaches, called \textit{Interpretable ML} (\textbf{IML}), are to train additional interpretable models to generate explanations, which usually are crucial %input 
features (i.e., important terms, in text analysis \citep{martens2013explaining, ribeiro2016should} or super-pixels, in image processing \citep{fong2017interpretable, selvaraju2017grad}), for each predicted outcome. However, most of existing IML algorithms usually treat input features independently, without considering their semantic correlations, especially in natural language processing. As a result, generated explanations commonly are fragmented, incomplete, and difficult to understand.

Addressing this problem is a non-trivial task, since: \textbf{(1)} It is difficult to capture semantic correlations among features, which can be contextually rich and dynamic; \textbf{(2)} There is still a lack of scientific study on how to integrate semantic correlations among features into IML to generate \textit{semantic} explanations, which are concise, complete, and easy to understand; and \textbf{(3)} The search space for semantic explanations can be large and complicated, given noisy and poor data. That results in a limited understanding of how to define semantic explanations, and effectively and efficiently identify them.
%and of how to effectively and efficiently identify them.

In literature, ontology, which encodes domain knowledge, can be used to capture semantic correlations among input features, such as entities, terms, phrases, concepts, etc. \citep{phan2017ontology, confalonieri2019ontology}. However, there is an unexplored gap regarding how to guide the learning process of an IML model based on ontology. Straightforwardly matching ontology and explaining data points, by randomly sampling co-occurring terms and concepts in conventional approaches, e.g., LIME \citep{ribeiro2016should}, may not generate semantic explanations, since contextual information in the data is usually rich  and complicated compared with the ontology. In addition, building an ontology that can sufficiently capture contextual information in the data is costly. Meanwhile, the traditional concept of anchor texts \citep{ribeiro2018anchors} can be used to narrow down the search space, by pinpointing generally important texts. However, the approach was not designed for each single and independent data point, i.e., at local level.

% \textbf{Our contributions.} To synergistically overcome these challenging issues, we propose a novel \textit{Ontology-based IML} (\textbf{OnML}), to generate good explanations. 
\textbf{Our contributions.} To synergistically overcome these challenging issues, we propose a novel \textit{Ontology-based IML} (\textbf{OnML}) to generate semantic explanations, by intergrating domain knowledge encoded in ontology and information extraction techniques into IML. 
In this paper, we consider a text classification model, in which text data is classified into different categories. Then, we learn a linear interpretable model by approximating the predictive model based on data sampled around the prediction outcome. 

%The key ideas of OnML are to \textbf{(1)} incorporate semantic correlations among words, which are captured in ontology, into learning interpretable representation, \textbf{(2)} learn anchor texts to reduce search space, and \textbf{(3)} combine ontology-based terms, anchor texts, and triples to generate good explanations.

In order to achieve our goals, we first present a new concept of \textit{ontology-based tuples}, each of which essentially is a set of correlated terms, words, and concepts semantically encoded and co-existed in the ontology and textual data. Departing from existing approaches, we identify and integrate ontology-based tuples into a new sampling approach, in which the semantic correlations among terms, words, and concepts are sampled and captured, instead of utilizing each of them independently. % term, word, and concept independently.

Second, we propose a new concept of \textit{learnable anchor texts}, to narrow down the search space for explanations. A learnable anchor text essentially is a contextual phrase, which can be expanded by adding nearby terms. %Anchors are chosen based on the model outcome without affecting their impacts to the model outcome.
% A learnable anchor text essentially is a contextual phrase, which can be expanded by adding nearby terms, without affecting the impact of the `anchor' to the model outcome. 
For instance, anchors can be started with a predefined seed term having negative meanings, e.g., ``no," ``not," ``illegal," and then be expanded to neighboring texts in order to effectively capture negative experiences and events, e.g., ``not get any help." Anchors, which have the highest \textit{importance scores} measuring their impacts upon the model outcome, will be chosen. %{\color{red} Triplexes consisting of subjects, predicate, and objects are further extracted from the text.}

Third, we introduce \textit{a set of regulations} to combine ontology-based tuples, anchor texts, and triplexes extracted from the text, to generate semantic explanations. Each  explanation is assigned an importance score. To our knowledge, OnML establishes the first connection among \textit{domain knowledge ontology}, \textit{IML}, and \textit{learnable anchor texts}. Such a mechanism will greatly extend the applicability of ML, by fortifying the models in both interpretability and trustworthiness.

Finally, extensive experiments conducted on two real-word datasets in critical applications, including drug abuse in the Twitter-sphere \citep{DBLP:journals/corr/abs-1904-02062} and consumer complaint analysis\footnote{\url{https://www.consumerfinance.gov/data-research/consumer-complaints/}}, to quantitatively and qualitatively evaluate our OnML, show that our algorithm generates concise, complete, and easy-to-understand explanations, compared with existing mechanisms. %\vspace{-10pt}

\section{Background and Problem Definition}

In this section, we revisit IML, ontology-based approaches, and information extraction algorithms, which are often used to generate  explanations. We further discuss the relation to previous frameworks and introduce our problem definition. 

Let $D$ be a database that consists of $N$ samples, each of which is a sample $x \in \mathbb{R}^d$ associated with its label $y$.
% \in \mathbb{Y}_K$, with $K$ possible categorical outcomes. 
Each $y$ is a one-hot vector of $K$ categories $y = \{ y_1, y_2, \ldots, y_K \}$. A classifier outputs class scores $f: \mathbb{R}^d \rightarrow \mathbb{R}^K$that maps an input $x$ to a vector of scores $f(x) = \{ f_1(x), f_2(x), \ldots, f_K(x) \}$ s.t. $\forall k \in [1,K]: f_k(x) \in [0,1]$ and $\sum_{k=1}^K f_k(x) = 1$. The highest-score class is selected as the predicted label for the sample. By minimizing a loss function $\mathcal{L}(f(x), y)$ that penalizes a mismatching between the prediction $f(x)$ and the original value $y$, an optimal classifier is selected.

\textbf{Interpretable Machine Learning.}
 Let us briefly revisit IML, 
 %interpretable machine learning techniques, 
 starting with the definition of \textit{interpretable model}.  Given an interpretable model $g$, which provides insights and qualitative understanding about the prediction model $f$ given an input $x$, there are two important criteria in learning $g$: 1) local fidelity, which implies the ability of interpretable models to approximate the prediction model in a vicinity of the input, and 2) interpretability, which is the sufficiently low complexity of interpretable models that make humans easy to understand the explanations. In textual data, the complexity, denoted as $T(g)$, usually is the number of important words \citep{martens2013explaining, ribeiro2016should}, based upon that users can easily handle to evaluate the generated explanations.

Let $z$ be a sample of $x$, where $z$ is generated by randomly selecting or removing features/words in $x$. $\phi_x(z)$ is a similarity function to measure the proximity between $x$ and $z$. Given a $d'$-dimensional binary vector $z' \in \{0,1\}^{d'}$, $z'_i = 1$ indicates that the feature $i$-th $(\in x)$ is present in $z$, and vice-versa. 
 
To achieve the interpretability and local fidelity, Ribeiro et al. \citep{ribeiro2016should} minimize a loss function  $L (f,g,\phi_x)$, with a low complexity $T(g)$, by solving the following problem: 
\begin{equation}\label{eq1}
    g^* = \arg\min_{g} L (f,g,\phi_x) + T(g)
\end{equation}
 where $ L (f,g,\phi_x) = \sum_z \phi_x(z) (f(z) - g(z'))^2$, $\phi_x(z) = \exp (-D(x,z)^2/\sigma^2)$ is an exponential kernel with $D(x,z)$ is a distance function (e.g., cosine distance for textual data) with a width $\sigma$, and  $g(z') = w_gz'$. %while considering $G$ consists of linear models. 
 
 To obtain the data $z$ for learning $g$ in Eq.~\ref{eq1}, sampling approaches are employed. In LIME \citep{ribeiro2016should}, the authors draw nonzero elements of the original data $x$ uniformly at random. Similar to this approach, a number of works follow \citep{nagrecha2017mooc,adhikari2018example,jia2019improving}. Apart from the randomization, model decomposition is another line of learning $g$ \citep{robnik2008explaining, martens2013explaining}, in which the prediction $f(x)$ is decomposed on individual features to learn the effect of each feature on the outcome.
 These existing randomization and decomposition approaches treat features independently; therefore, they cannot capture correlations among features. This may not be practical in real-world scenarios, since features usually are highly and semantically correlated. %, i.e., text data. 

\textbf{Ontology-based Approaches.}
 To capture semantic correlations among input features and ontology %(i.e., consisting of related concepts and their relations) 
 can be applied. 
Ontology is used in \citep{freddy2018semantic} to filter and rank concepts from selected data points to conduct informative  explanations. The explanations are derived in ontological forms. For example, the information, ``a 30 year-old individual, with an operation occurred in 1989,'' can be conveyed by the representation, ``\textit{TheSilentGeneration $\sqcap$ OperationIn1980s.}'' (TheSilentGeneration denotes people in the age range of 30-39.) 
 However, building a rich contextual ontology is expensive, so typically ontology only captures a limited number of core concepts and their correlations. This is the reason why ontological forms cannot capture all common sense knowledge in the textual information. In reality, humans generally use natural languages in a variety of text presentations. Therefore, an appropriate combination of a single-form ontology with other approaches to generate semantic explanations is necessary.

%\textit{TheSilentGeneration $\sqcap$ OperationIn1960s} which can be used to represent the information: 30 year-old individual (TheSilentGeneration denotes people in the age 30s) with an operation occurred in 1964. 

 \begin{figure}[t]
      \centering
      \includegraphics[scale=0.34]{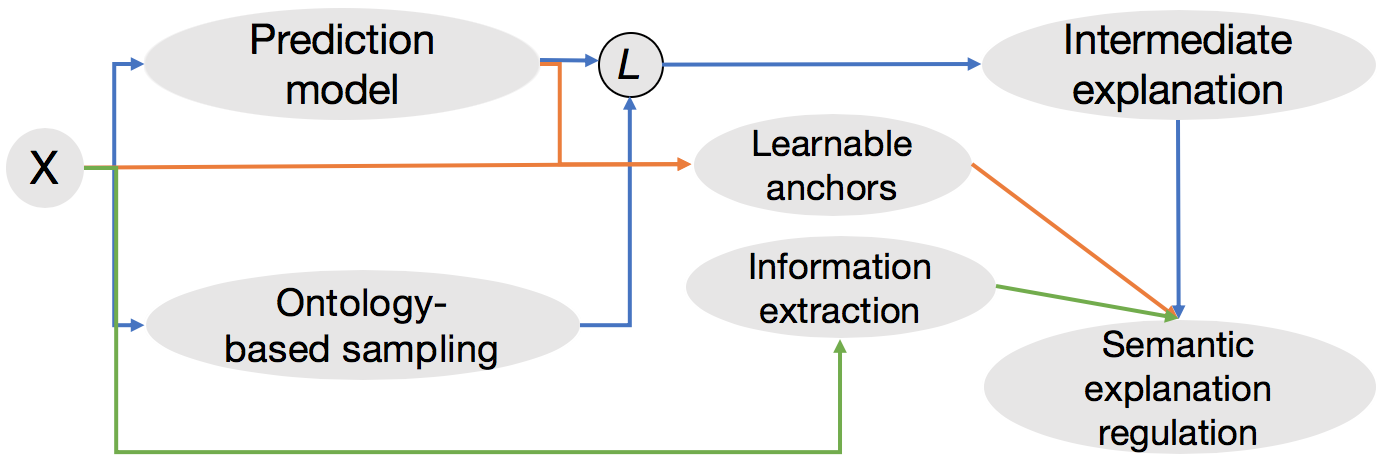} %\vspace{-5pt}
      \caption{A flow chart of the OnML approach.} %\vspace{-10pt}
      \label{fig1}
 \end{figure}
 
  \begin{figure}[t]
      \centering
      \includegraphics[scale=0.45]{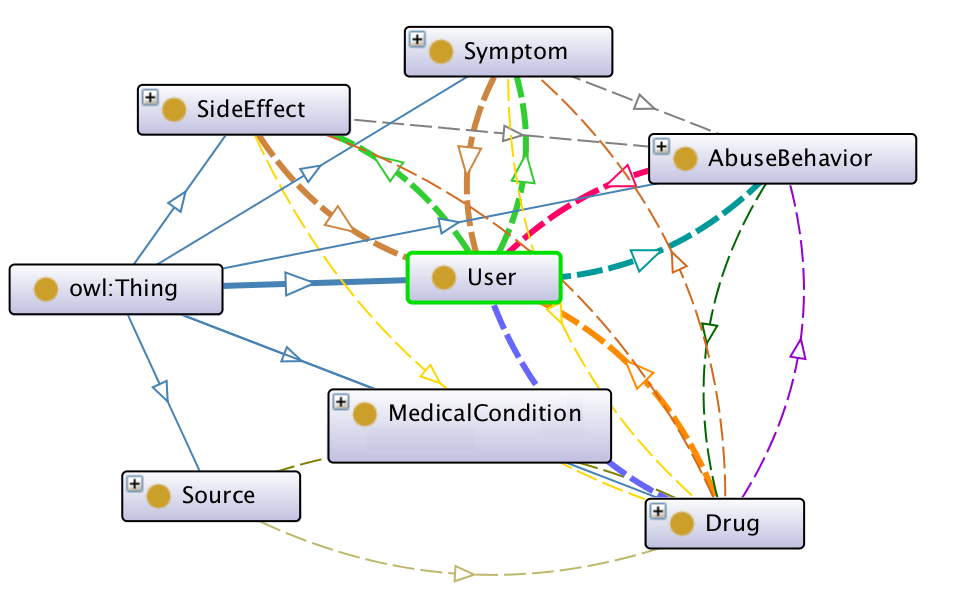} %\vspace{-5pt}
      \caption{Drug abuse ontology.} %\vspace{-15pt}
      \label{fig2}
 \end{figure}

In \citep{confalonieri2019ontology}, Confalonieri et al. use ontology to learn an understandable decision tree, which is an approximation of a neural network classifier. %The explanation in this framework is generated in a form of reading features learned by the decision tree. 
 Explanations are in a non-syntactic form, and they are not designed to explain a single and independent data point. Different from \citep{confalonieri2019ontology}, we aim at generating semantic explanations for each input $x$. 
In this paper, generating semantic explanations is defined as a process of mapping a text to a representation of important information in \textit{a syntactic} and \textit{understandable} form.

\textbf{Information Extraction.}
Apart from IML, information extraction (IE) is another direction to capture contextual information semantically. The first Open IE algorithm is TextRunner \citep{banko2007open}, which identifies arbitrary relation phrases in English sentences by automatically labeling data using heuristics for training the extractor. 
Following \citep{banko2007open}, a number of Open IE  \citep{wu2010open, soderland2010adapting, fader2011identifying} were introduced. Unfortunately, these approaches ignore the context. OLLIE \citep{schmitz2012open} includes contextual information; and extracts relations mediated by nouns, adjectives, and verbs; and outputs triplexes (subject, predicate, object). Compared to Open IE approaches, our algorithm mainly focuses on generating semantic explanations associated with the prediction label.

\section{Ontology-based Interpretable Machine Learning for Textual Data}
In this section, we formally present our proposed OnML framework (Fig.~\ref{fig1}). Alg.~\ref{alg1} presents the main steps of our approach. Given an input $x$, an ontology $\mathcal{O}$, and a set of all concepts $\mathcal{C}$ in $\mathcal{O}$, we first present the notion of \textit{ontology-based tuples} (Line 3), which will be used in an \textit{ontology-based sampling technique} to learn the interpretable model $g$ (Lines 4-6). Next, we learn potential anchor texts using the input $x$ and the model $f(x)$ (Line 7). Meanwhile, OLIIE \citep{schmitz2012open}~is applied to extract triplexes, which have high confident scores, in $x$ (Line 8). After learning $g$, learning anchor texts $\mathcal{A}$, and extracting triplexes $\mathcal{T}$, we introduce a set of regulations to combine them together to generate semantic explanations (Line 9). Let us first present the notion of ontology-based tuples as follows.

%which is simply two concepts which correlations are captures by the ontology.

 \subsection{Ontology-based Tuples}
 
Given concepts $A$ and $B$, $A \mapsto B$ is used to indicate that $A$ has a directed connection to $B$. In considerably correlated domains, such as text data, it is observed that 1) words appeared near to each other in a sentence have the same contextual information, and 2) different sentences usually have different contextual information. To encode the observations, we introduce a \textit{contextual constraint}, as follows:%\vspace{-5pt}
 \begin{equation}\label{eq2}
    \lambda_{x_k}(x_l) \le \gamma
\end{equation}
where $x_k$ and $x_l$ are two words in $x$,  $\gamma$ is a predefined threshold, and $\lambda_{x_k}(x_l)$ measures the distance between the positions of $x_k$ and $x_l$ in $x$. In text data, if $x_k$ and $x_l$ belong to two sentences, they are considered to be violating the contextual constraint. Intuitively, the constraint is used to connect words 1) that appear near to each other in a sentence (contextual correlated) and 2) that belong to connected concepts in the ontology (conceptual correlated). If there is no contextual constraint, there can be mismatched information between the domain knowledge and the explanation extracted in the text.

\begin{comment} For instance, given a consumer complaint ontology (Fig.~\ref{fig2}) and $x$ as ``They send me a message that my application is denied since I did not pay fee.", the list of ontology-based words can be found \{send, application, denied, pay, fee\}. If we only use conceptual knowledge from the ontology, it may result in 

By using $\gamma$ (e.g., $\gamma = 4$), we can eliminate ``send" from the explanation, if not, it may create an explanation as ``send application" that is contextually incorrect for this complaint.
\end{comment}

%For instance, in the text $x$ ``She does not like  weed. Smoke causes addiction and headache.", if $\gamma = 4$ then (smoke, addiction) with $\lambda_{\text{smoke}}(\text{addiction}) = 2$ satisfies the contextual constraint. Meanwhile, (weed, Smoke) does not satisfy the constraint, since they are in different sentences. 
%The formal definition of an ontology-based tuple is as follows:
\begin{mydef}\label{def1} Ontology-based tuple. Given $x_k$ and $x_l$ in $x$, $(x_k, x_l)$ is called an ontology-based tuple, if and only if: (1) $\exists A, B \in \mathcal{C}$ s.t. $x_k \in A$ and $x_l \in B$; (2) $A \mapsto B$; and (3) $\lambda_{x_k}(x_l) \le \gamma$.
\end{mydef}%\vspace{-5pt}

Since ontology has directed connections among its concepts, ontology-based tuples are asymmetric, i.e., $(x_k, x_l)$ and $(x_l, x_k)$ are different. 
For the sake of clarity without affecting the generality of the approach, we use a drug abuse ontology as an example (Fig.~\ref{fig2}). %The drug abuse ontology was built by our domain experts. 
Given the drug abuse ontology and $x$ as \textit{``She  uses orange juice and does not like weed. She knows that smoke causes addiction and headache."},  list of ontology-based words can be found \{use, weed, smoke, addiction, headache\}. These words are in ``Abuse Behavior" (use and smoke), ``Drug" (weed), ``Side Effect" (addiction), and ``Symptom" (headache) concepts. Following the  aforementioned conditions (Eq. \ref{eq2} with $\gamma= 3$), two ontology-based tuples are found, which are (smoke, addiction) and (smoke, headache). In the meantime, (addiction, headache) and (weed, smoke) are not ontology-based tuples, since there is no directed connection between the ``Side Effect" concept and the ``Symptom" concept, and ``weed" and ``smoke" are in different sentences. By using the contextual constraint, we can eliminate ``use weed," which is contextually incorrect, from the explanation.

\begin{algorithm}[t] 
\small
\caption{OnML approach}\label{alg1}
\begin{algorithmic}[1]
\STATE \textbf{Input:} Input $x$; ontology $\mathcal{O}$, and user-predefined anchor $\mathcal{A}_0$
\STATE Classify $x$ by a prediction model $ f: \mathbb{R}^d \rightarrow \mathbb{R}^K $
\STATE Find ontology-based tuples $(x_i, x_j)$ in $x$ based on concepts and relations in $\mathcal{O}$
\STATE Sample $x$, based on ontology-based tuples found by our sampling technique to obtain sampled data $z \in \mathcal{Z}$
\STATE Generate vectors of predictive scores $f(z)$ with $z \in \mathcal{Z}$
\STATE Learn an interpretable model $g$ based on $f(z)$ and $g(z')$ by Eq.~\ref{eq1}\\
\STATE Learn anchor text by our anchor learning algorithm (Alg.~\ref{alg2})
\STATE Extract triplexes in $x$ using an existing Open IE technique
\STATE Combine ontology-based tuples, learned anchors, and extracted triplexes by our proposed regulations
\STATE \textbf{Output:} Semantic explanation $\mathcal{E}$  
\end{algorithmic}
\end{algorithm} %\vspace{-5pt} 

%to find $g^*$ we need to minimize the locality loss function $L (f,g,\phi_x)$ with respect to a low complexity $T(g)$. I
 %Unlike LIME (\citealt{ribeiro2016should}) that randomly and independently samples terms/words in $x$, 

 \subsection{Ontology-based Sampling Technique}%\vspace{-5pt} 
To integrate ontology-based tuples  into learning $g$, we introduce a novel ontology-based sampling technique. To learn the local behavior of $f$ in its vicinity (Eq.~\ref{eq1}), we approximate $L (f,g,\phi_x)$ by drawing samples based on $x$, with the proximity indicated by $\phi_x$. %We sample $x$ given the extracted ontology-based tuples, as follows.
 A sample $z$ can be sampled as: %\vspace{-5pt}
\begin{equation}\label{eq3}
    z = \Big( \cup_{x_i \in x, i \neq k, i \neq l} \mathcal{R} (x_i) \Big) \cup \mathcal{R}(\{x_k, x_l \}) %\vspace{-5pt}
\end{equation}
where $\mathcal{R}(x_i)$ and $\mathcal{R}(\{x_k, x_l \})$ are probabilities randomly drawn for each word $x_i \in x (i \neq k, l)$ and words $x_k, x_l \in x$ together, respectively. If $\mathcal{R}$ is greater than a predefined threshold, then the word(s) will be included in $z$. 
% where $\mathcal{R}(\alpha)$ is a probability randomly drawn for each word $\alpha$ in $x$. If $\mathcal{R}(\alpha)$ is greater than a pre-defined threshold, then $\alpha$ will be included in $z$.
 
In our sampling process, $x_k$ and $x_l$, i.e., an ontology-based tuple, are sampled together as a single element. 
This aims to integrate the semantic correlation between $x_k$ and $x_l$, captured in an ontology-based tuple into the sampling process. In fact, we are sampling the semantic correlation, but not sampling each word/feature $x_k$ or $x_l$ independently. This enables us to measure the impact of this semantic correlation on $f(x)$. In addition, words, which are not in any ontology-based tuple, are sampled independently.
After sampling $x$ (Eq.~\ref{eq3}), we obtain the dataset $\mathcal{Z}$ that consists of sampled data points $z$ associated with its label $f(z)$. 
$\mathcal{Z}$ is used to learn $g^*$ by solving Eq.~\ref{eq1}. %\vspace{-1.5pt}

%optimize the loss function $L(f, g , \phi_x)$ in Eq.~\ref{eq1} to find the optimal $g^*$. 

%The binary indicator vector $z'$ is further obtained based on the appearance of each word in $z$ compared to its appearance in $x$.

%that satisfy Eq.~\ref{eq2} as a single word and uses the learned classifier to determine their contribution to the classification.

 \subsection{Learnable Anchor Text} %\vspace{-1.5pt}
 
 Before presenting our anchor mechanism, we introduce an \textit{importance score} notion, which will be used to choose the best anchor and calculate the importance of generated explanations.

 \subsubsection{Importance Score}
 To get insights into the importance of generated explanations and their impacts upon the model outcome, we calculate an importance score ($IC$) for each explanation. Intuitively, the higher importance score, the more important the explanation is. $IC$ is calculated as: %\vspace{-5pt}
    \begin{equation}\label{eq5}
    % \small
        IC (r) = \bar{c}_r \Big( f(x) - f(x/r) \Big)  %\vspace{-5pt}
    \end{equation}
    where $x/r$ is the original text $x$ excluding words in the explanation $r$ and $\bar{c}_r$ is average coefficients of $g^*$ associated with all words in $r$. 
    % $f(\cdot)$ is a vector of the predictive scores,
\begin{algorithm}[t] 
\small
\caption{Anchors learning algorithm}\label{alg2}
\begin{algorithmic}[1]
\STATE \textbf{Input:} Input $x$; prediction model $f$; number of sentences in $x$, denoted as $M$; user-predefined anchors $\mathcal{A}_0$
\STATE $\mathcal{A} \leftarrow \emptyset  ~ (\mathcal{A}:$ set of anchors for $x$)
\FOR {$i \in M $}
\IF {any $\mathcal{A}_0$ appears in the sentence $i$}
\STATE Denote $D_{\mathcal{A}}$ as a set of ordered words appearing after $\mathcal{A}_0$ in the sentence $i$ in $x$
\STATE $\mathcal{A}_n \leftarrow \emptyset $ ($\mathcal{A}_n$ is a set of candidate anchors)
\STATE $\mathcal{F}_n \leftarrow \emptyset $ ($\mathcal{F}_n$ is a set of importance scores, associated with each candidate anchor)
\FOR {$x_j \in D_{\mathcal{A}}$}
\STATE $\mathcal{A}_n \leftarrow \mathcal{A}_0 \cup x_j$; $\mathcal{A}_0 \leftarrow \mathcal{A}_n$; $\mathcal{F}_n \leftarrow \mathcal{F}_n \cup IC(\mathcal{A}_n)$
\ENDFOR
\STATE \quad Choose the best anchor for sentence $i$:  $\mathcal{A}_i = \arg\max_{\mathcal{A}_n}  \mathcal{F}_n$%\vspace{-5pt}
% \begin{equation}
%     \mathcal{A}_i = \arg\max_{\mathcal{A}_n}  \mathcal{F}_n \nonumber
% \end{equation} \vspace{-15pt}
\ELSE 
\STATE $\mathcal{A}_i \leftarrow \emptyset $
\ENDIF
\ENDFOR
\STATE \quad $\mathcal{A}  \leftarrow \mathcal{A} \cup \mathcal{A}_i  $
\STATE \textbf{Output:}  $\mathcal{A}$ 
\end{algorithmic}
\end{algorithm}

 \subsubsection{Anchor Text Learning Mechanism}
 %We consider how to deal with long and poor data, which consists of a large number of words, misspelled text, or informal writing, etc., in explaining text classification. 
 It is challenging to work with long and poor data, e.g., large number of words, or misspelled text, since the contextual information  is generally rich and complicated. Building an ontology to adequately represent such data is expensive, and insufficient in many cases. That results in a large undercovered search space for  explanations. To address this problem, we introduce a learnable anchor mechanism to narrow down the search space.

The learning anchor technique is presented in Alg.~\ref{alg2}. The anchor is initialized with an empty set (Line 2). A set of user-predefined anchors $\mathcal{A}_0$ is provided, which consists of starting-words that are further expanded by incrementally adding words to the end of the sentence. Then, the importance score of each candidate anchor is calculated, following Eq.~\ref{eq5}. The top-1 anchor $\mathcal{A}$, which has the highest important score, for each sentence are then chosen. %\vspace{-1.5pt}

%based on the maximum importance score associated with each starting-word. 

%In each sentence, for each word in  $\mathcal{A}_0$  that appears in the text, we learn an anchor text from a set of candidate anchors $\mathcal{A}_n$, which is created by incrementally adding succeeding words to the starting-word anchor towards the end of the sentence. 

 \subsection{Generating Semantic Explanations} %\vspace{-1.5pt}
 
We further apply OLLIE \citep{schmitz2012open} to extract triplexes  $\mathcal{T}$ (subject, predicate, and object)  to identify the syntactic structure in a sentence, which can shape our explanations in a readable form. To generate semantic explanations $\mathcal{E}$, we introduce a set of regulations to combine $g^*$, $\mathcal{A}$, and $\mathcal{T}$ together: \\
\textbf{1)} $\mathcal{E} \subseteq D_x$ with $D_x$ is a set of all words in $x$.\\
\textbf{2)} If there is no ontology-based tuple found, $\mathcal{E}$ will only consist of the learned anchor texts. \\
\textbf{3)} In a sentence, if there are two or more ontology-based tuples, we introduce four rules to merge them together:
    \begin{itemize}
    \item \textit{Simplification:}
        \begin{itemize}
            \item Given $(x_k, x_l)$ and $(x_k, x_m)$, if $x_l$ and $x_m$ are in the same concept, then the ontology-based explanation is $\{x_k,x_l \textit{ and/or } x_m\}$.
            \item Given $(x_k, x_m)$ and $(x_l, x_m)$, if $x_k$ and  $x_l$ are in the same concept, then  the ontology-based explanation is $\{x_k \textit{ and/or } x_l,x_m\}$.
            \item Given $(x_k, x_l)$ and $(x_l, x_m)$, then  the ontology-based explanation is $\{x_k,x_l,x_m\}$.
        \end{itemize}
        
        \item \textit{Union}: Given $(x_k, x_l)$, $(x_k, x_m)$, $(x_l, x_m)$, and $\{x_k, x_l, x_m\}$, the ontology-based explanation is $\{x_k,x_l,x_m \}$.
        
        \item \textit{Adding Causal words}: Semantic explanation can be in the form of a causal relation. Thus, if a causal word, e.g., ``because," ``since," ``therefore," ``while," ``whereas," ``thus," ``thereby," ``meanwhile, ``however," ``hence," ``otherwise," ``consequently," ``when," ``whenever" appears between any words in ontology-based tuples/explanations, we add the word to the explanation, following its position in $x$.
        \item \textit{Combining with anchor texts $\mathcal{A}$ and triplexes $\mathcal{T}$}: After having ontology-based explanations, we combine them with $\mathcal{A}$ and $\mathcal{T}$ based on their positions in $x$. Then, the \textit{semantic explanation} is generated from the beginning towards the end of all positions of words found in the ontology-based explanations, $\mathcal{A}$, and $\mathcal{T}$. For example, in the sentences, ``We were filling out all the forms in the application. However, there is a letter in saying loss mitigation application denied for not sending information to us.", after the learning process, we obtain: 1)  ontology-based explanation is (loss, application); 2) anchor text is ``not sending information;" and 3) triple is ``a letter; denied; mitigation application." The explanation $\mathcal{E}$ is ``a letter in saying loss mitigation application denied for not sending information."    \end{itemize}
\textbf{4)} If different ontology-based tuples are in different sentences in $x$, due to the contextual constraint in Eq.~\ref{eq2}, the explanation for each sentence follows the 3$^{rd}$ regulation.
    
 It is worthy noting that we use aforementioned regulations to combine ontology-based tuples to be a longer ontological term. This makes the ontology used in a much better representation rather than independent and direct connections $A \mapsto B$. %For instance, in the previous example, instead of having two ontology-based tuples (smoke, addiction) and (smoke, headache), we shorten them into (smoke, addiction and/or headache). 
    %%%%%%%%%%% Experiments %%%%%%%%%%% 
\begin{figure*}[t]
      \centering
      \includegraphics[scale=0.45]{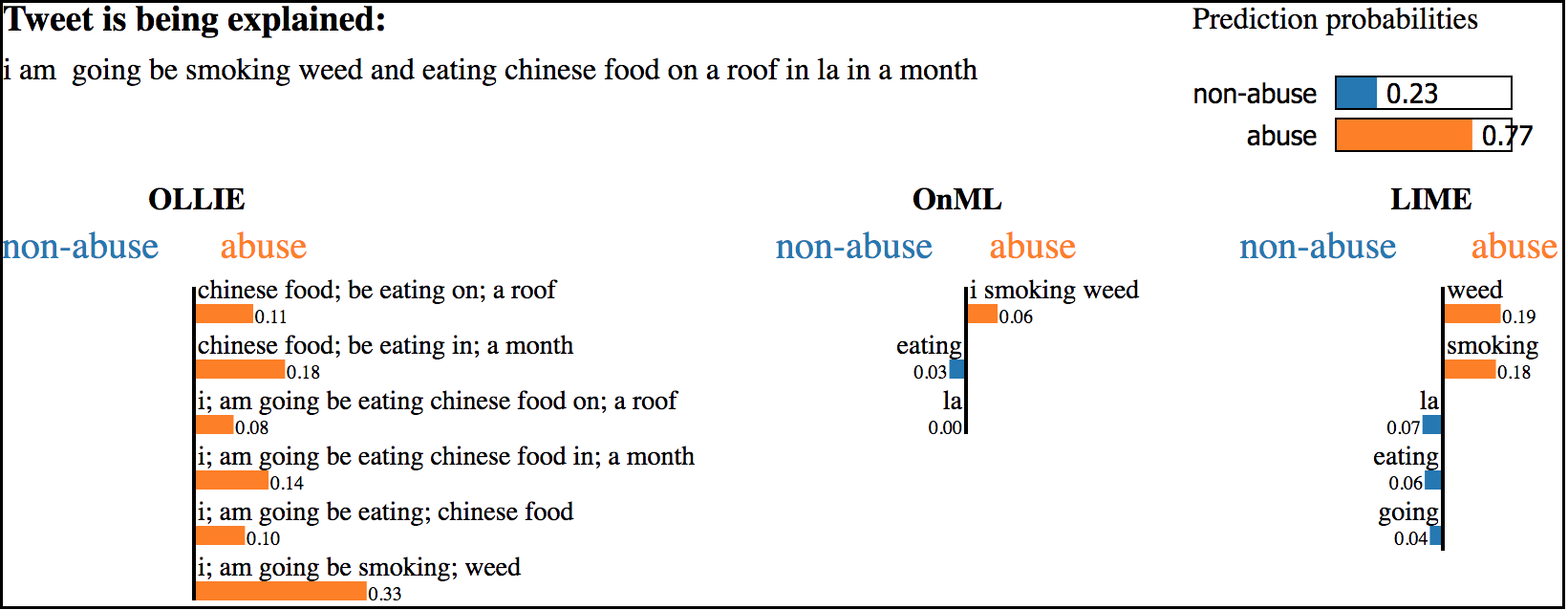} %\vspace{-5pt}
    %   \caption{Visualization of a drug abuse experiment.} 
   % \vspace{-12pt}
      \label{fig3a}
 \end{figure*}
 
 \begin{figure*}[t]
      \centering
      \includegraphics[scale=0.4]{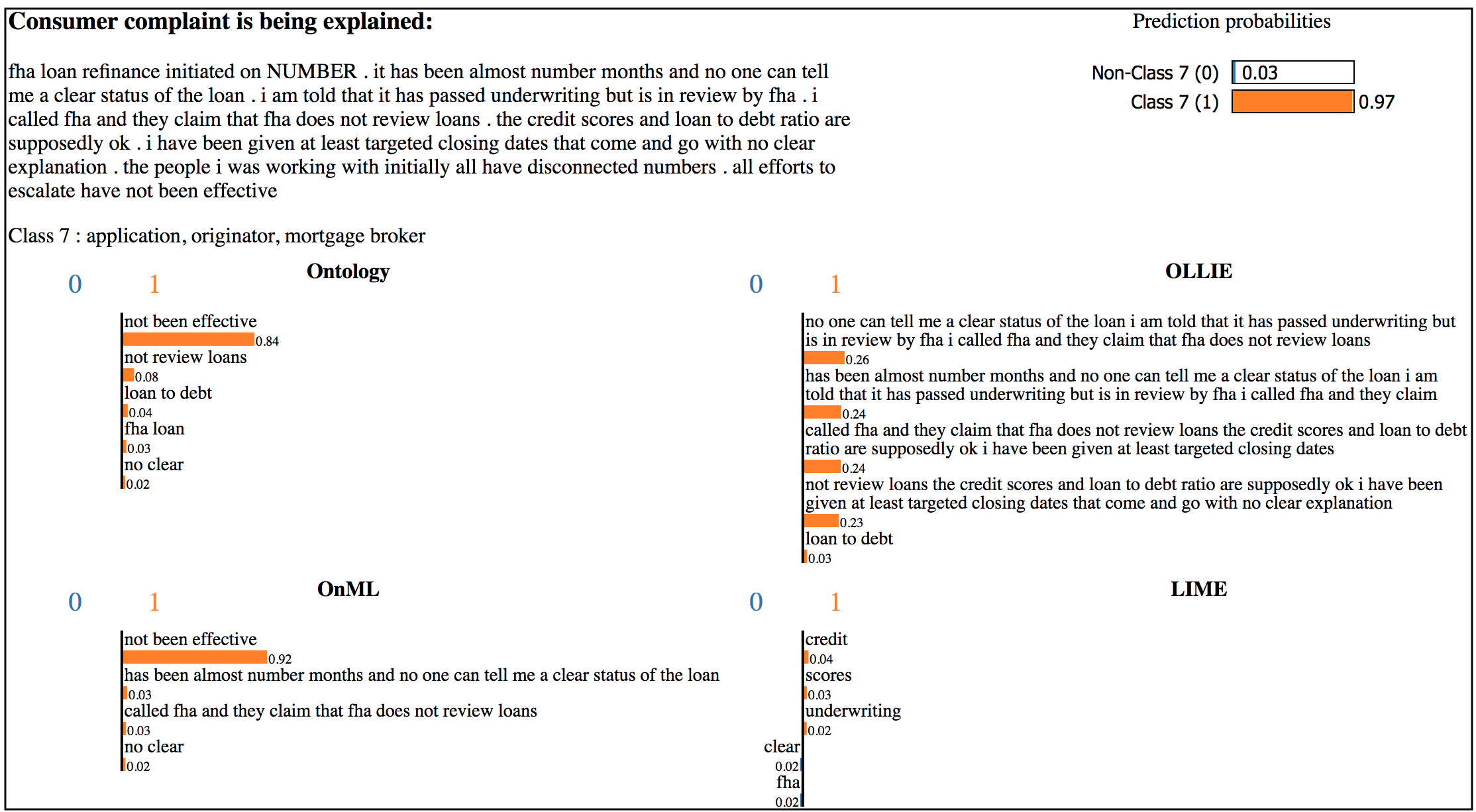} %\vspace{-15pt}
      \caption{Visualization of drug abuse (\textit{top}) and consumer complaint (\textit{bottom}) experiments.} %\vspace{-12.5pt}
      \label{fig3}
 \end{figure*}

\section{Experiment} %\vspace{-10pt}
We have conducted extensive experiments on two real-world datasets, including drug abuse (Twitter-sphere \citep{DBLP:journals/corr/abs-1904-02062}) and consumer complaint analysis from Consumer Financial Protection Bureau$^1$.

\subsection{Baseline Approaches} %\vspace{-2.5pt}
Our OnML approach is evaluated in comparison with traditional approaches: \textbf{(1)} an interpretable model-agnostic explanation, i.e., LIME \citep{ribeiro2016should}; and \textbf{(2)} information extraction, i.e., OLLIE \citep{schmitz2012open}. LIME is one of the state-of-the-art and well-applied approaches in IML, %interpretable model-agnostic explanation, 
in which the predictions of any model are explained in a local region near the sample being explained. There are other algorithms sharing the same spirit as LIME, in terms of generating explanations \citep{lundberg2017unified, selvaraju2017grad, shrikumar2017learning, sundararajan2016gradients, springenberg2014striving, bach2015pixel}.
%, such as SHAP \citep{lundberg2017unified}, GCam \citep{selvaraju2017grad}, DeepLIFT \citep{shrikumar2017learning}, Integrated Gradients \citep{sundararajan2016gradients}, Layerwise Relevance Propagation \citep{bach2015pixel}, Guided-Backpropagation \citep{springenberg2014striving} and Simonyan-Gradient \citep{simonyan2013deep}. 
For the sake of clarity, we use LIME as a representative baseline regarding this line of research. 

The key differences among OnML, OLLIE, and LIME  are that OnML leverages domain knowledge to tie the explanations up to the predicted label and considers correlations among words in textual data to generate semantic explanations. Meanwhile, OLLIE focuses more on grammatical analysis to extract triples %(subject, predicate, object) 
from the text. LIME generates fragmented interpretable components by learning a linear interpretable model locally around the prediction outcome and weight these components using coefficients of the interpretable model. In LIME and OLLIE, domain knowledge is not used.

%\vspace{-2.5pt}

\subsection{Datasets and Domain Ontologies}% \vspace{-2.5pt}
To validate the proposed method, we have developed two different domain ontologies, which are drug abuse ontology (Fig.~\ref{fig2}) and consumer complaint ontology (in the \textbf{Appendix})
% \footnote{\url{https://www.dropbox.com/s/4v2lnv68vxs4dvg/Appendix.pdf?dl=0}}). 
These ontologies were constructed for certain  domains (e.g., drug abuse and consumer complaint) since it is necessary to capture specific semantic and causal relations among components. As default in Prot{\'e}g{\'e} \citep{knublauch2004protege},
each arrow represented by its color demonstrates a certain type of causal relation in which its tail represents a domain and its head represents a range of the relation. For example, in the drug abuse ontology (Fig.\ref{fig2}), purple arrow is for ``is involved with" with domain is ``Drug" and range is ``Abuse Behavior" while green arrows are for ``suffer from". These ontologies were semi-manually generated, in which concepts were grouped and collected from the dataset by K-means clustering algorithm \citep{forgy1965cluster}, and then judged by humans to reduce inappropriate concepts.

% \vspace{-5pt}

%A more detailed description of our ontologies is available in the Appendix (supplementary file).

\subsubsection{Drug Abuse Dataset}

We will use the term ``drug abuse" in the wider sense, including abuse and use of Schedule 1 drugs that are illegal and have no medical use (e.g. legal painkiller and weed) or illegally (e.g. getting drugs without prescription or even from blackmarket); and misuse of Schedule 2 drugs, which have medical uses, yet have a potential for severe addiction, and which can be life-threatening \citep{barlas2013prescription}.  The drug abuse ontology captures different concepts collected from drug abuse tweets, grouped by K-means clustering algorithm, and then finalized by our team experts.
Main concepts of the drug abuse ontology (\textbf{DrugAO}) (Fig.~\ref{fig2}) capture correlation among key concepts, including abuse behaviors, drug types, drug sources, drug users, symptoms,  side effects, and medical condition when using drug.  Abuse behaviors concept is about behaviors of abusers, such as abuse, addict, blunt, etc. Drug types consists of different types of legal and illegal drugs, e.g., narcotics, cocaine, and weed. Drug sources is where drug users, who are the main objects of the ontology, gets drugs from. Symptoms and side effects are about different negative short-term and long-term effects of drugs on users. Medical condition contains terms about expression of disease and illness caused by using drugs. In total, DrugAO has $506$ drug-abuse related terms (including slang terms and street names), and $18$ relations.

The drug abuse dataset (Table~\ref{tb1}) consists of $9,700$ tweets labelled by \citep{DBLP:journals/corr/abs-1904-02062} with a high agreement score. Among them, $3,043$ tweets are drug abuse tweets, labeled \textit{positive}  and the rest are non drug abuse tweets, labeled \textit{negative}. %These tweets were used for learning a ML model to predict whether a tweet is drug-abuse-related or not (described in the next section).

% \begin{table}
% \caption{Data statistical analysis.} \vspace{-5pt}
% % \scriptsize
% % \small
% \label{tb1}
% \begin{center}
% % \begin{small}
% \begin{tabular}{|l|c|c|}
% \hline
% % \begin{small}
% \diagbox[width=12em]{Statistics}{Dataset} & Drug abuse & \makecell{Consumer \\ complaint} \\
% % \end{small}
% \hline \hline
% $\#$ of samples & 9,700 & 13,965 \\
% $\#$ of categories & 2 & 16\\
% Max $\#$ of words/sentence & 37 & 4,893 \\
% Mean $\#$ of words/sentence & 12 & 285\\
%  \hline
% \end{tabular} \vspace{-10pt}
% % \end{small}
% \end{center}
% \end{table} 
% % \vspace{-10pt}

% \begin{table}
% \caption{AC and SC in drug abuse.} \vspace{-5pt}
% \label{tb2}
% \begin{center}
% % \begin{small}
% \begin{tabular}{| l | c | c|}
% \hline
%  & Accuracy changes (\%) & Score changes (\%) \\
% \hline \hline
% LIME  & 15.04 & 26.98 \\
% OLLIE  & 15.47 & 23.52  \\
% \textbf{OnML} & \textbf{25.52} & \textbf{33.48}  \\
%  \hline
% \end{tabular} \vspace{-10pt}
% % \end{small}
% \end{center}
% \end{table} 

\subsubsection{Consumer Complaint Dataset} 
% A consumer complaint is defined, here, as a complaint about a range of consumer financial products and services, sent to companies for response. Main concepts of the consumer complaint ontology (\textbf{ConsO}) encode the relation among different entities related to consumer complaint: for instance, who is complaining; what happened to make consumers unhappy and then complaint; etc. 
% In total, we have $572$ finance and product-related terms and $9$ relations covered in our ontology.
% The consumer complaint dataset consists of $13,965$ mortgage-related complaints, labeled with $16$ categories. 
% These complaints were used for learning a model to predict the issue regarding each complaint.

\begin{figure}[t]
      \centering
      \includegraphics[scale=0.5]{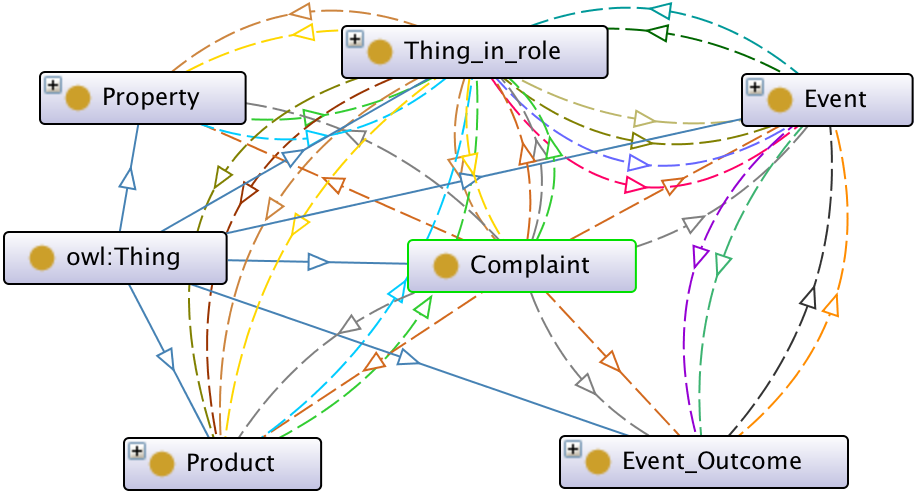}
      \caption{Consumer complaint ontology.}
      \label{fig7}
 \end{figure}

  A consumer complaint is defined, here, as a complaint about a range of consumer financial products and services, sent to companies for response.  In complaints, consumers typically talk about their mortgage-related issues, such as: (1) Applying for a mortgage or refinancing an existing mortgage (application, credit decision, underwriting); (2) Closing on a mortgage (closing process, confusing or missing disclosures, cost); (3) Trouble during payment process (loan servicing, payment processing, escrow accounts); (4) Struggling to pay mortgage (loan modification, behind on payments, foreclosure); (5) Problem with credit report or credit score; (6) Problem with fraud alerts or security freezes, credit monitoring or identity theft protection services; and (7) Incorrect information on consumer's report or improper use of consumer's report.  Main concepts of the consumer complaint ontology (\textbf{ConsO}) (Fig.~\ref{fig7}) encode the relation among different entities related to consumer complaint: for instance, who is complaining; what happened to make consumers unhappy and then complaint; etc. There are six major concepts in ConsO, which are thing in role, complaint, event, event outcome, property, and product. Thing in role is people and organizations related to complaint, such as buyers, investors, dealers, et,. Event and event outcome are about negative events happened that cause consumer complaints. Property is things belonging to consumers and product is substances of some parties (e.g., banks) offering to consumers. 
In total, we have $572$ finance and product-related terms and $9$ relations covered in our ontology.
The consumer complaint dataset consists of $13,965$ mortgage-related complaints, labeled with $16$ categories. 
These complaints were used for learning a model to predict the issue regarding each complaint.

% \vspace{-2.5pt}

\subsection{Experimental Settings}

% \vspace{-2.5pt}

Our experiment focuses on validating whether: \textbf{(1)} Our OnML approach can be applied on different agnostic predictive models; and \textbf{(2)} Our approach can generate better explanations, compared with baseline approaches, in both quantitative and qualitative measures. Our ontologies, code, and data are available on Github\footnote{\url{https://github.com/PhungLai728/OnML}}.

To achieve our goal, we carry out our evaluation through three approaches. First, by employing SVM and LSTM, we aim to illustrate that OnML works well with different agnostic predictive models. Second, we leverage the word deleting approach \citep{arras2017relevant} as an quantitative evaluation. Third, we apply qualitative evaluation with Amazon Mechanical Turk (\textbf{AMT}).

\subsubsection{Model Configurations}
 In the drug abuse dataset, tweets were vectorized by TF-IDF \citep{ramos2003using} and then classified by a linear kernel SVM model. We achieved $83.6\%$ accuracy. Tweets are short, i.e., the average and maximum numbers of words in a tweet are 12 and 37 (Table~\ref{tb1}). Therefore, it is not necessary to apply the anchor learning algorithm, which is designed to tighten down the search space for long text data. 

 In the consumer complaint dataset, Word2vec \citep{mikolov2015computing} is applied for feature vectorization. Then, a Long short-term memory (LSTM) \citep{hochreiter1997long} is trained as a prediction model. In LSTM, we used an embedding input layer with $d=300$, one hidden layer of $64$ hidden neurons, and a softmax output layer with $16$ outputs. An efficient ADAM \citep{kingma2014adam} optimization algorithm with learning rate $0.01$ was employed to train LSTM.
For the prediction model, we achieved $53\%$ accuracy. We registered that this is a reliable performance, since the $16$ categories are densely correlated resulting in a lower prediction accuracy \citep{deng2010does}. Another reason for the low accuracy is the limited number of samples. We will collect more data in the future.

%In fact, LSTM is complex enough to capture the contextually rich and dynamic characteristics of the consumer complaint dataset; meanwhile, SVM is effective in classifying drug abuse tweets. 

For sufficiently learning anchors in consumer complaints, we have chosen a set of negative terms as user-predefined anchors $\mathcal{A}_0 = \{$not, no, illegal, against, without$\}$. %In addition, in LIME, the maximum number of explanation features is set to $5$. 
Importance scores in LIME are weights of the linear interpretable model. With OLLIE, importance scores of extracted triplexes are calculated in the same way as in our method (as shown in Eq.~\ref{eq5}).  LIME and OLLIE settings are used as default in \citep{ribeiro2016should, schmitz2012open}. We only show OLLIE rules which have the confidence score greater than $0.7$ and top-$5$ words from LIME. The contextual constraint $\gamma$ in Eq.~\ref{eq2} is $3$ for drug abuse and $10$ for consumer complaint dataset. The pre-defined threshold in Eq.~\ref{eq3} is $0.5$.

\textbf{To be fair}, we also combined the learned anchors to the results of OLLIE. In addition, another variation of our algorithm is to combine ontology-based terms and anchors, called \textbf{Ontology} algorithm. This is further used to comprehensively evaluate our proposed approach.

\subsubsection{Quantitative Evaluation}
 %To conduct a quantitative experiment, 
We use the word deleting approach \citep{arras2017relevant}, which deletes a sequence of words from a text and then re-classifies the text with missing words. By differences between the original text and the missing text, we examine the importance of the explanation to the prediction.
 %Intuitively, if an explanation is important, removing the explanation will deteriorate the prediction result. In that case, it will reduce the accuracy and change the prediction score $f(x)$ significantly. 
 Accuracy changes (AC) and prediction score changes (SC) are as:%\vspace{-4pt}
 \begin{align}
%  \small
\text{AC} & = \text{Original accuracy} - \frac{ \sum_{i = 1}^{|test|} \text{Updating accuracy}}{|test|} \nonumber \\ %\vspace{-10pt} 
\text{SC} &= \frac{ \sum_{i = 1}^{|test|} IC (\text{top-$k$ explanations of $i$-th sample}) }{|test|} \nonumber
\end{align}%\vspace{-5pt} 
where the higher values of AC and SC indicate the more important explanations derived. 

\begin{table}
\caption{Data statistical analysis.} %\vspace{-5pt}
% \scriptsize
% \small
\label{tb1}
\begin{center}
% \begin{small}
\begin{tabular}{|l|c|c|}
\hline
% \begin{small}
\diagbox[width=12em]{Statistics}{Dataset} & Drug abuse & \makecell{Consumer \\ complaint} \\
% \end{small}
\hline \hline
$\#$ of samples & 9,700 & 13,965 \\
$\#$ of categories & 2 & 16\\
Max $\#$ of words/sentence & 37 & 4,893 \\
Mean $\#$ of words/sentence & 12 & 285\\
 \hline
\end{tabular} %\vspace{-10pt}
% \end{small}
\end{center}
\end{table} 
% \vspace{-10pt}

\begin{table}
\caption{AC and SC in drug abuse.} %\vspace{-5pt}
\label{tb2}
\begin{center}
% \begin{small}
\begin{tabular}{| l | c | c|}
\hline
 & Accuracy changes (\%) & Score changes (\%) \\
\hline \hline
LIME  & 15.04 & 26.98 \\
OLLIE  & 15.47 & 23.52  \\
\textbf{OnML} & \textbf{25.52} & \textbf{33.48}  \\
 \hline
\end{tabular} %\vspace{-10pt}
% \end{small}
\end{center}
\end{table}

 In our experiment, we deleted the top-$k$ highest importance score explanations in OnML and OLLIE approaches and the top-$m$ highest weighted words in LIME. To be fair, $m$ is the number of words in the $k$-deleted explanations in OnML.  
 In drug abuse, $k = 1$ 
 since the tweet is typically short, and so there are not many explanations generated. In consumer complaint classifying, $k \in \{1, 2, 3\}$.

  \begin{figure}[t]
%   \small
      \centering
      \includegraphics[scale=0.6]{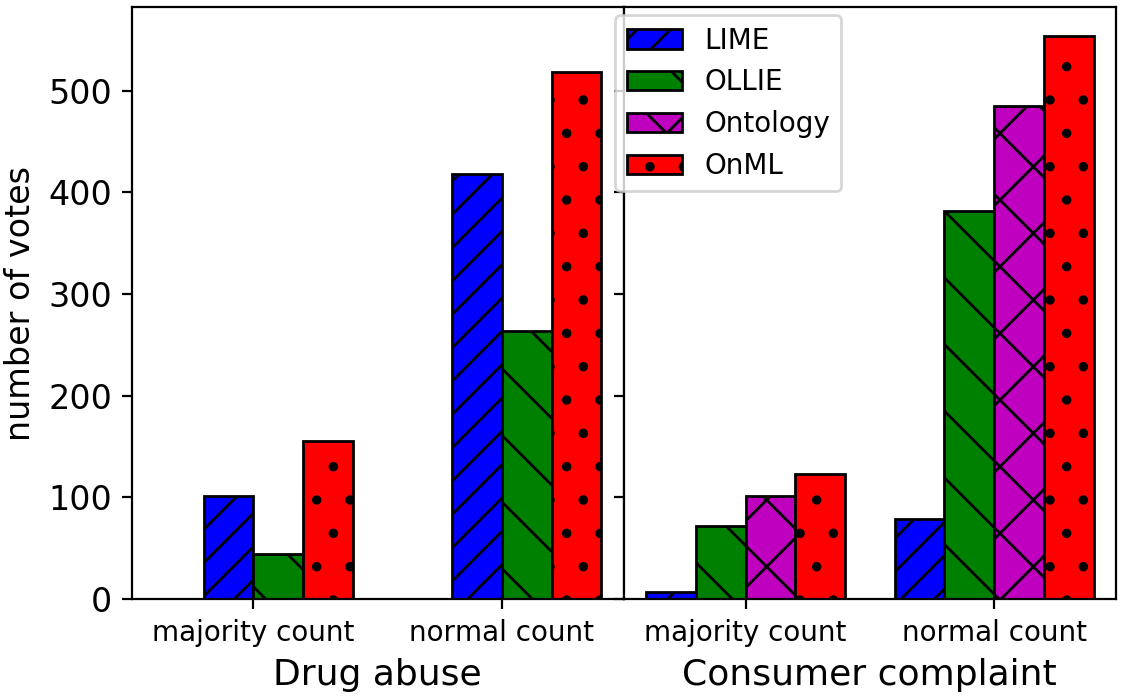} %\vspace{-5pt}
      \caption{AMT experiment results.} %\vspace{-15pt}
      %\vspace{-10pt}
      \label{fig5}
 \end{figure}
 
  \begin{figure}[t]
      \centering
      \includegraphics[scale=0.18]{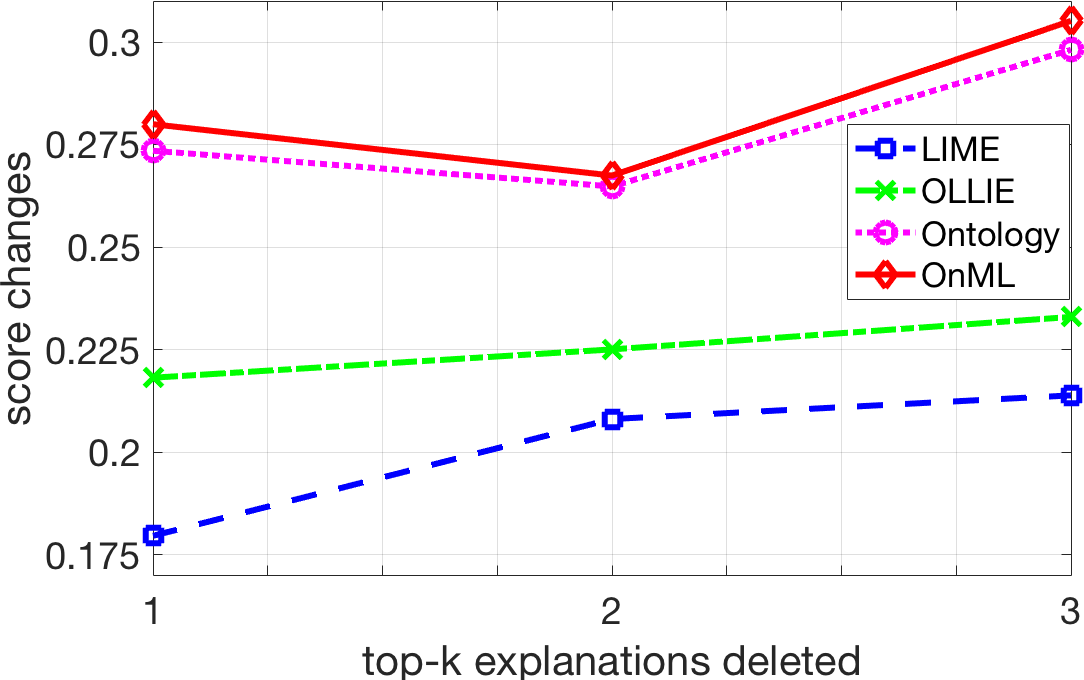} %\vspace{-5pt}
      \caption{Average score changes in consumer complaint.} %\vspace{-10pt}
      \label{fig16}
 \end{figure}

 \subsubsection{Qualitative Evaluation}
We recruit human subjects on Amazon Mechanical Turk (AMT). %, in which machine learning experts are not required. 
This is a common means of evaluation for the needs of qualitative investigation by humans %, since semantic explanations judged by machine are inadequate
\citep{martens2007comprehensible, selvaraju2017grad}.  Detailed guidance for each experiment is provided to users before they conduct the task.

We asked AMT workers to choose the best explanation by seeing side-by-side explanation algorithms. On top of that, we provided the original tweet/ complaint associated with their labels and  prediction results. The visualization showing explanation results of the approaches is in Fig.~\ref{fig3}. \textbf{It is important to note that}, in our real experiment, to avoid bias, name of each algorithm is hidden, and their positions in the visualization are randomized.

We were recruiting $4$ users/tweets in the drug abuse and $5$ users/complaints in the consumer complaint experiment.  
%There are two ways to quantify the voting results from AMT users: 
To quantify the voting results from AMT users, we use:
\textbf{(1)} Count the total number of votes, called \textit{normal count}, i.e., the best algorithm is chosen over all $1,500$ votes ($5$ users/complaint $\times$ $300$ complaints); and \textbf{(2)} Count the majority number of votes, called \textit{majority count}, i.e., the best algorithm for each complaint is the algorithm of the largest number  over $5$ votes.  
% \vspace{-5pt}

\subsection{Experimental Results and Analysis} 
%\vspace{-5pt}
To evaluate the interpretability of each approach, $300$ positive tweets and $300$ complaints, randomly selected, were used. 
% In this paper, $300$ positive tweets and $300$ complaints, randomly selected, were used to evaluate the interpretability of each approach. 
% \vspace{-10pt}

\subsubsection{Drug Abuse Explanation}
As in Table~\ref{tb2},  %shows the AC and SC values of each algorithm. 
the accuracy is deducted significantly, and the predictive score changes the most in OnML. In fact, the values of AC and SC are 25.52\% and 33.48\% given OnML, compared with 15.47\% and 23.52\% given OLIIE, and 15.04\% and 26.98\% given LIME. This demonstrates that the explanations generated by our algorithm are more significant, compared with the ones generated by baseline approaches. In the evaluation by humans using AMT (Fig.~\ref{fig3}), OnML clearly outperforms LIME and OLLIE. Text in the tweet is generally short and can be represented by several key words. Therefore, individual words learned by LIME can be sufficient to generate more insightful explanations. Meanwhile, OLLIE tends to extract all possible triplexes in the text, which can be redundant and wordy explanations.

\subsubsection{Consumer Complaint Explanation}
The results on the consumer complain dataset further strengthen our results. Fig.~\ref{fig16} shows SC after deleting top-$1$, top-$2$, and top-$3$ explanations from OnML, Ontology, and OLLIE, as well as after deleting the most important words in LIME. In all three cases, score changes in OnML have the highest values, indicating that the explanations generated by OnML are the most significant to the prediction. In the evaluation by humans using AMT (Fig.~\ref{fig3}), our OnML algorithm outperforms baseline approaches. Ontology approach achieves higher results than LIME and OLLIE. This shows the effectiveness of the ontology-based approach. LIME does not consider semantic correlations among words, resulting in a poor outcome. % in both quantitative and human subject experiments. %LIME generated word-by-word explanations. %\vspace{-5pt}

\subsubsection{Completeness and Concision}
In Fig.~\ref{fig3} (\textit{top}), OnML generates ``i smoking weed," which provides concise and complete information about why it is predicted as a drug abuse tweet (smoking weed) and who was doing it (i) in a syntactic form. Meanwhile, 1) LIME derives relevant words to drug abuse (i.e., weed, smoking) without considering the correlation among these words; and 2) OLLIE generates lengthy and somewhat irrelevant explanations, e.g., ``chinese food; be eating on; a roof." In Fig.~\ref{fig3} (\textit{bottom}), OnML derived semantic explanations for consumer complaints, which tell us that consumers were facing issues in loan refinance, e.g., ``called fha and they clain that fha does not review loans." Compared to OnML, Ontology generates laconic explanations, e.g., ``fha loan" that give no sense of why consumer complaints. LIME provides a set of fragmented words and OLLIE generates wordy explanations, which are difficult to follow. More examples of explanation results are in the \textbf{Appendix}.
%More examples of drug abuse and consumer complaint explanation are in the \textbf{Appendix}$^5$.

Our key observations are: \textbf{(1)} Combining ontology-based tuples, learnable anchor texts, and information extraction can generate complete, concise, and insightful explanations to interpret the prediction model $f$; and \textbf{(2)} Our OnML model outperforms other baseline approaches in both the quantitative and qualitative experiments, showing a promising result.

\section{Conclusion}   %\vspace{-10pt}
In this paper, we proposed a novel ontology-based IML to generate semantic explanations, by integrating  interpretable models, ontologies, and information extraction techniques. A new ontology-based sampling technique was introduced, to encode semantic correlations among features/terms in learning interpretable representations. An anchor learning algorithm was designed to limit the search space of semantic explanations. Then, a set of regulations for connecting learned ontology-based tuples, anchor texts, and extracted triplexes is introduced, to produce semantic explanations. Our approach achieves a better performance, in terms of semantic explanations, compared with baseline approaches, illustrating a better interpretability into ML models and data. Our approach paves an early brick on a new road towards gaining insights into machine learning using domain knowledge.

% OnML achieved some promising results. However, many scopes for improvements can be considered for future work. One of potential directions is to explore the generality of OnML by leveraging more different knowledge bases and data, e.g., WikiData \citep{vrandevcic2014wikidata}, DBPedia \citep{auer2007dbpedia}, and WikiDB \citep{mehler2008wikidb}, in which uncertainty and bias could be critical issues. Second, a reliable semi-supervised approach to construct ontology with theoretical guarantees regarding the consistency between domain knowledge and textual data is urgently needed in order to apply our OnML at scale given different domain applications.

\textbf{Acknowledgment.} The authors gratefully acknowledge the support from the National Science Foundation (NSF) grants CNS-1747798, CNS-1850094, and NSF Center for Big Learning (Oregon). We thank for the financial support from Wells Fargo. We also thank Nisansa de Silva (University of Oregon, USA) for valuable discussions.

\bibliography{IEEEexample}
\bibliographystyle{IEEEtran}

\appendix

% \textbf{Domain Ontologies}

%  \begin{figure}[ht]
%       \centering
%       \includegraphics[scale=0.47]{figure/drug_ontology.png}
%       \caption{Main concepts of a drug abuse ontology generated by our team experts.}
%       \label{fig4}
%  \end{figure}

%   \begin{figure}[h]
%       \centering
%       \includegraphics[scale=0.5]{figure/cc_onto2.png}
%       \caption{Consumer complaint ontology.}
%       \label{fig7}
%  \end{figure}
%   \vspace{-10pt}
 
%  , for instance who is complaining, what happened to make consumers unhappy and then complaint, etc.. 

% \textbf{Consumer Complaint Ontology and Additional Experimental Results}\\
The following figures are additional experiment results for drug abuse and consumer complaint experiment. %Further experiment results for consumer complaint are provided on Github.
 \begin{figure}[h]
      \centering
      \includegraphics[scale=0.3]{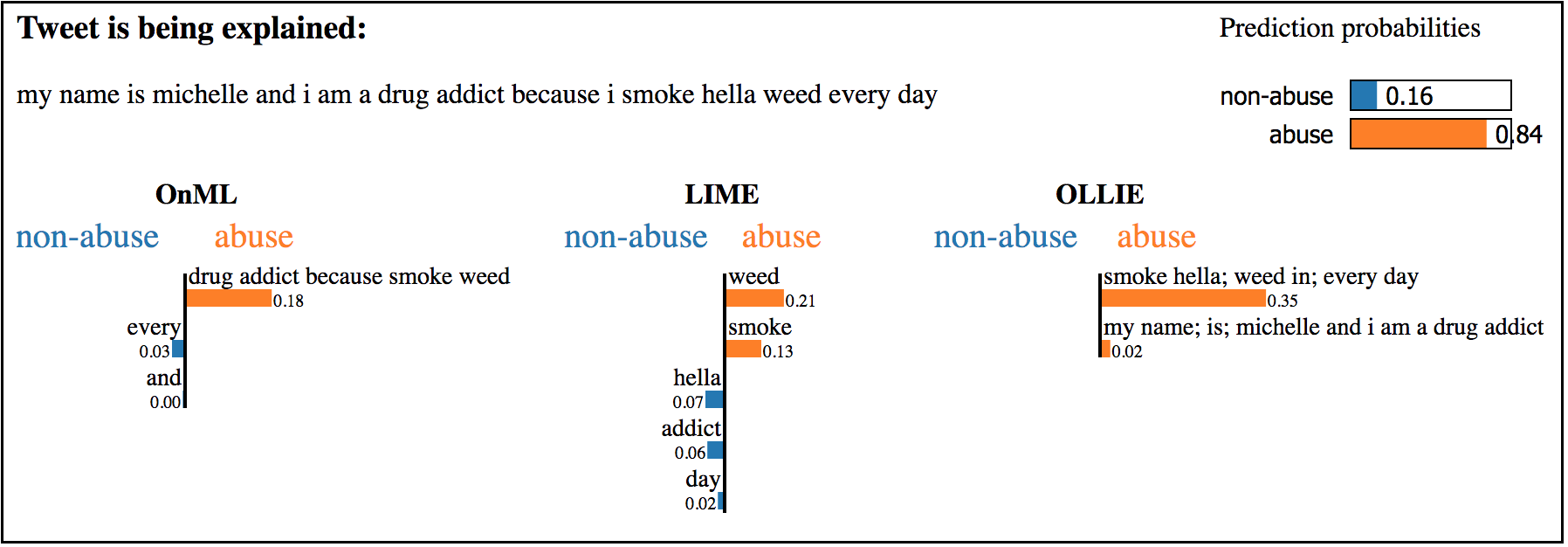}
    %   \caption{Visualization of a drug abuse experiment. }
      \label{fig6}
 \end{figure}
 \vspace{-25pt}

%  \begin{figure*}[ht]
%       \centering
%       \includegraphics[scale=0.49]{figure/tw6291.png}
%       \caption{Visualization of a drug abuse experiment. }
%       \label{fig7}
%  \end{figure*}

 \begin{figure}[h]
      \centering
      \includegraphics[scale=0.285]{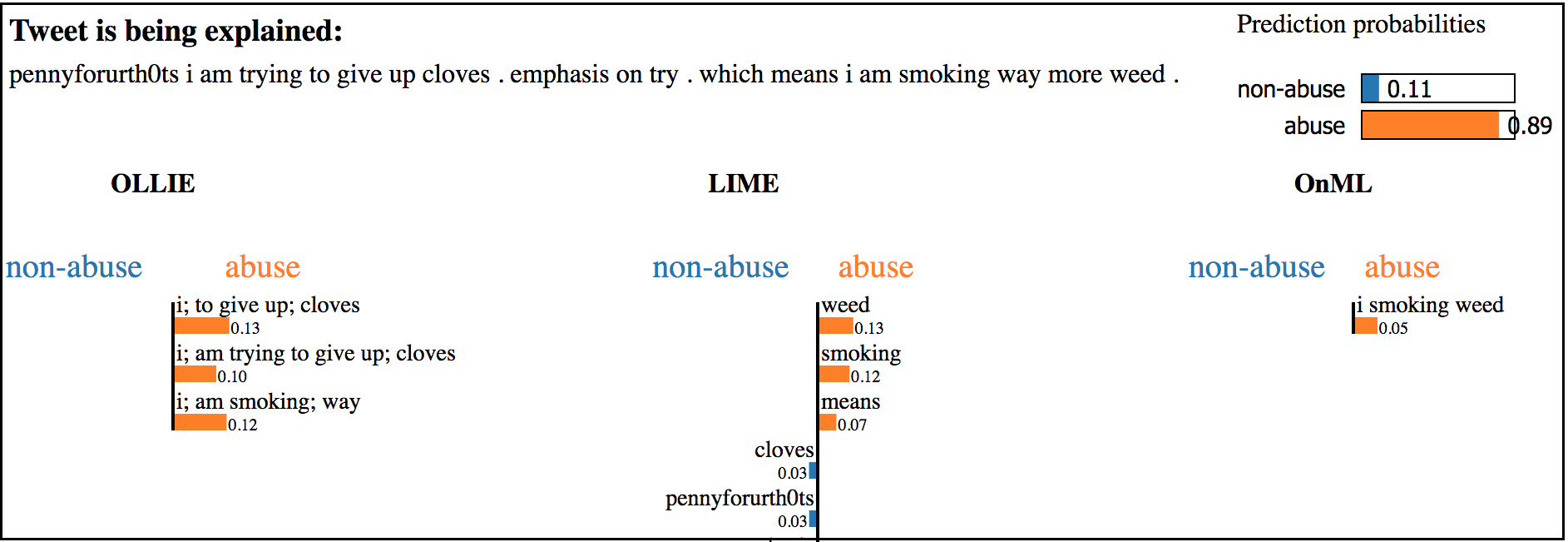}
    %   \caption{Visualization of a drug abuse experiment.}
      \label{fig8}
 \end{figure}
  \vspace{-25pt}

 \begin{figure}[h]
      \centering
      \includegraphics[scale=0.295]{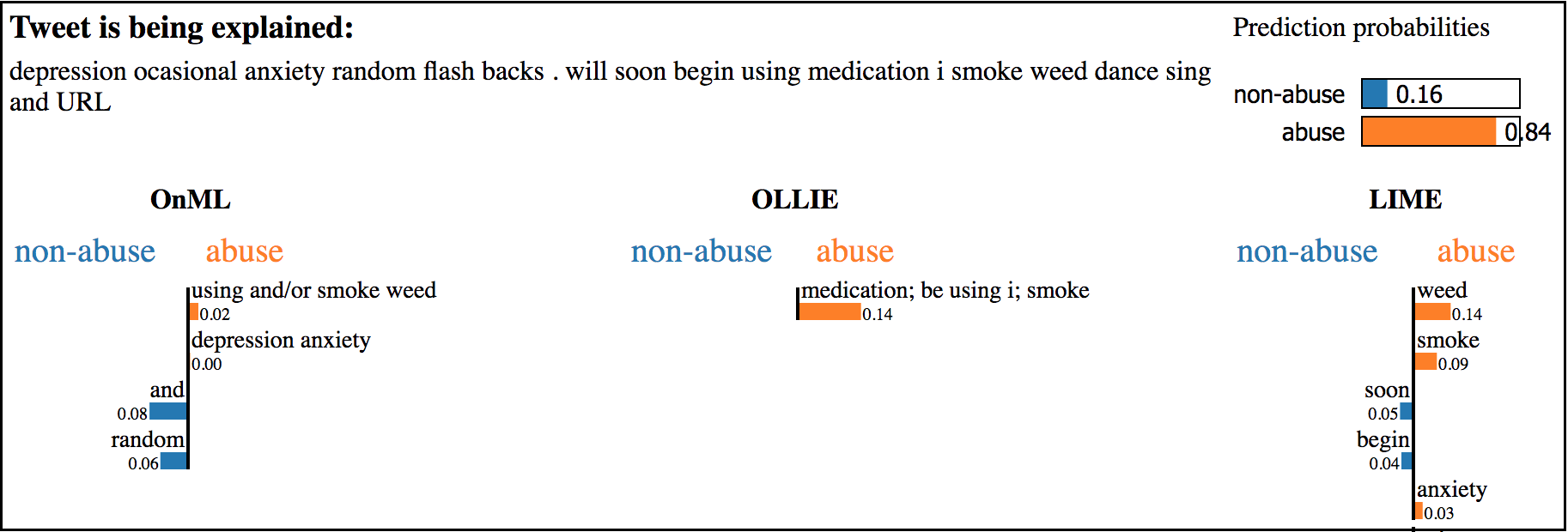}
    %   \caption{Visualization of a drug abuse experiment. }
      \label{fig9}
 \end{figure}
  \vspace{-25pt}

 \begin{figure}[hbt!]
      \centering
      \includegraphics[scale=0.355]{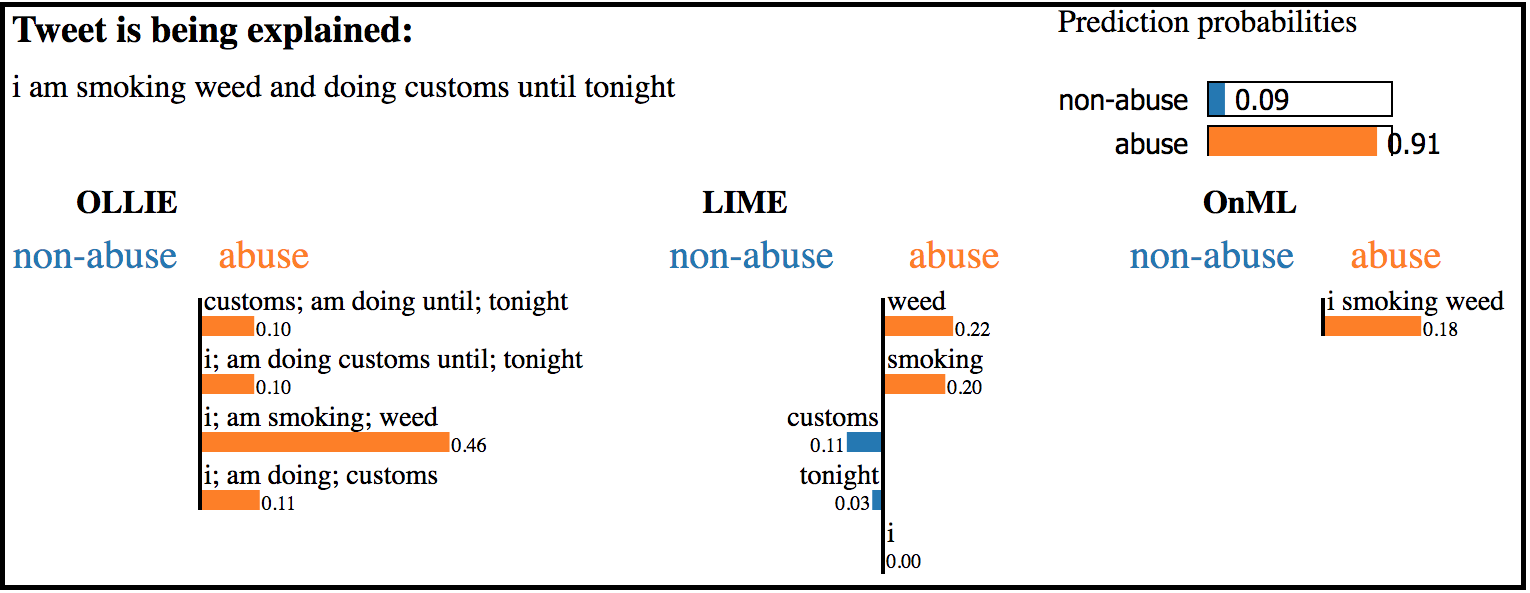}
    %   \caption{Visualization of a drug abuse experiment. }
      \label{fig10}
 \end{figure}
%   \vspace{-30pt}

 \begin{figure*}[h]
      \centering
      \includegraphics[scale=0.43]{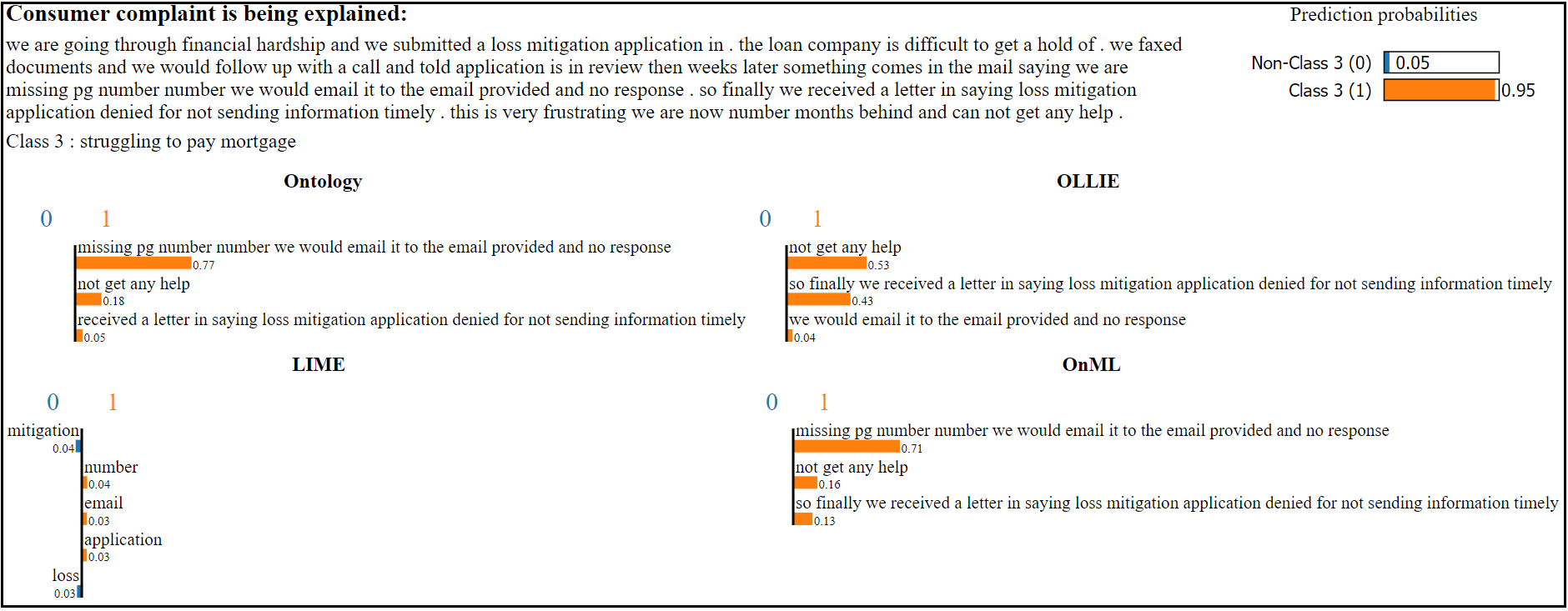}
    %   \caption{Visualization of a consumer complaint experiment. }
      \label{fig11}
 \end{figure*}
  %\vspace{-10pt}
  
% %  \begin{figure*}[ht]
% %       \centering
% %       \includegraphics[scale=0.36]{figure/cc136.PNG}
% %       \caption{Visualization of a consumer complaint experiment. }
% %       \label{fig12}
% %  \end{figure*}

 \begin{figure*}[h]
      \centering
      \includegraphics[scale=0.37]{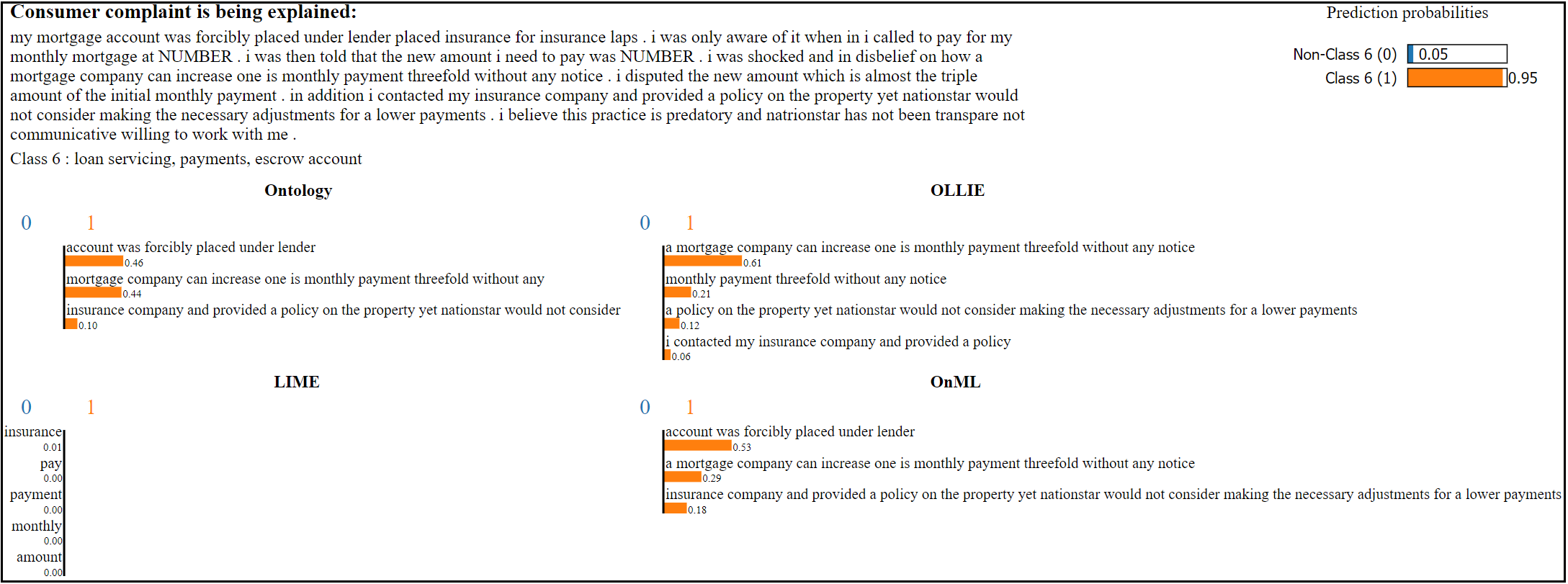}
    %   \caption{Visualization of a consumer complaint experiment. }
      \label{fig13}
 \end{figure*}
  %\vspace{-10pt}
  
 \begin{figure*}[h]
      \centering
      \includegraphics[scale=0.31]{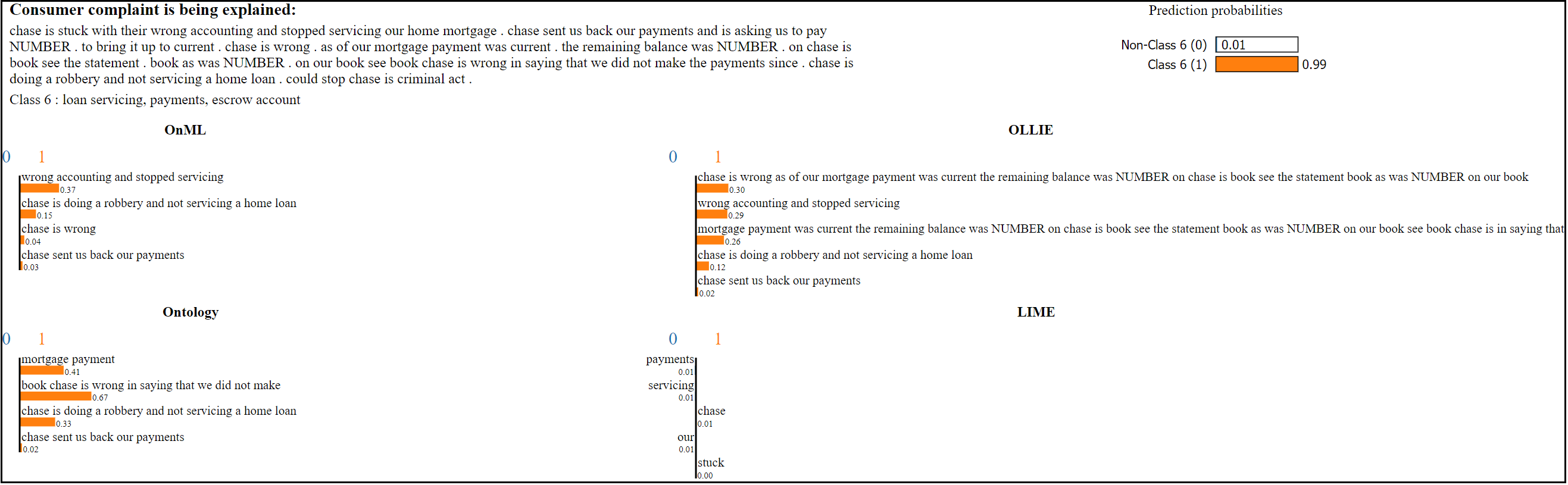}
    %   \caption{Visualization of a consumer complaint experiment.}
      \label{fig14}
 \end{figure*}
  %\vspace{-10pt}
  
 \begin{figure*}[h]
      \centering
      \includegraphics[scale=0.445]{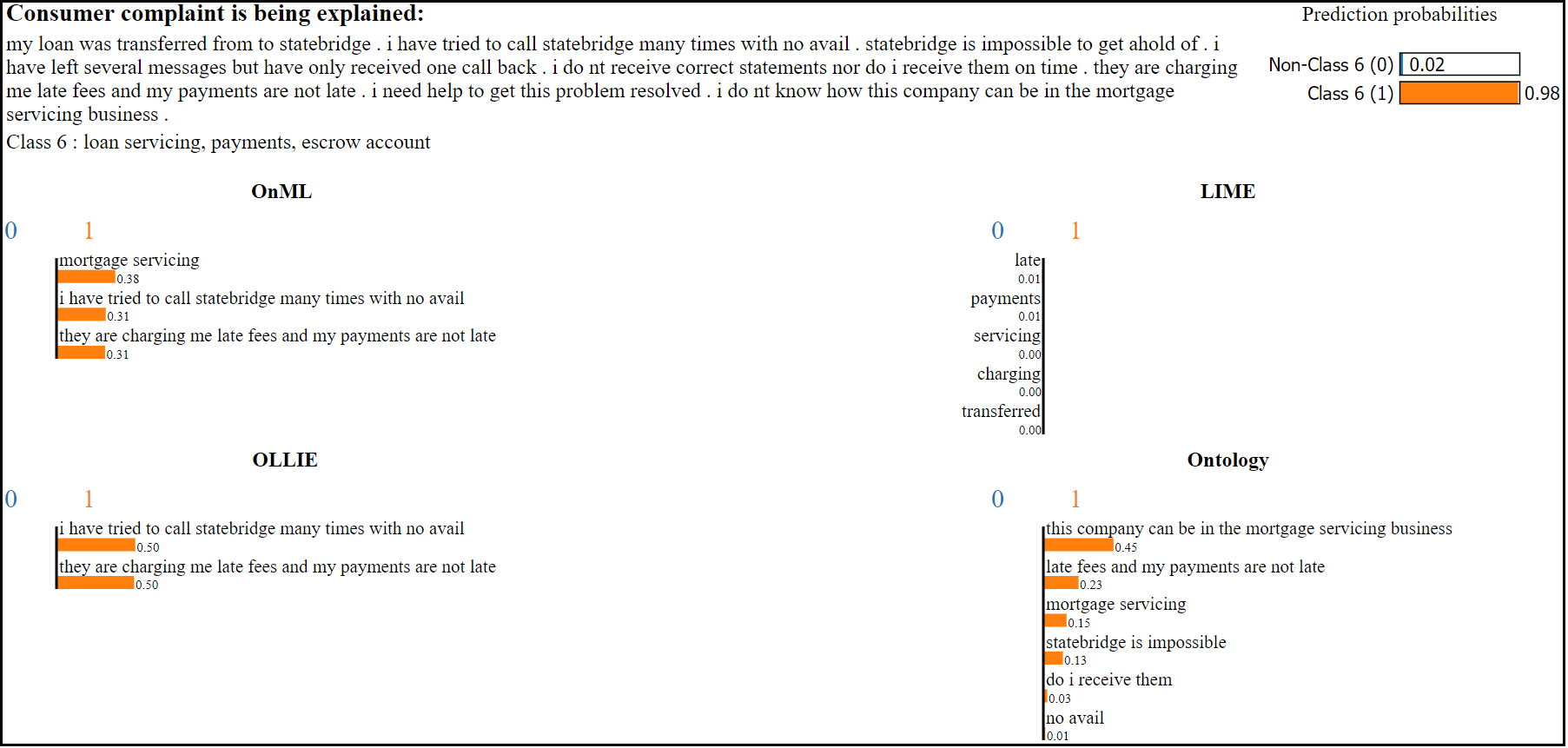}
    %   \caption{Visualization of a consumer complaint experiment. }
      \label{fig15}
 \end{figure*}
 % \vspace{-10pt}

\end{document}